**Daily Dose Paper (2979 words)**

**Title:**

The Daily Dose: Workflow-Integrated Large Language Model Automation for Clinical Summarization and Trial Identification in Radiation Oncology

**Authors:**

***[1]Jason Holmes, Ph.D. – holmes.jason@mayo.edu,**
***[2,3,4]Federico Mastroleo, M.D. - mastroleo.federico@mayo.edu,**
**[2]Mariana Borras-Osorio, M.D. - borras-osorio.mariana@mayo.edu,**
**[2]Srinivas Seetamsetty - seetamsetty.srinivas@mayo.edu,**
**[2]Satomi Shiraishi, Ph.D. - shiraishi.satomi@mayo.edu,**
**[1]Mirek Fatyga, Ph.D. - fatyga.mirek@mayo.edu,**
**[5]Judy C. Boughey, M.D. - boughey.judy@mayo.edu,**
**[5]Cornelius A. Thiels, D.O. - thiels.cornelius@mayo.edu,**
**[2]William G.Breen, M.D. - breen.william@mayo.edu,**
**[2]Daniel J. Ma, M.D. - ma.daniel@mayo.edu,**
**[2]Daniel K. Ebner, M.D. - ebner.daniel@mayo.edu,**
**[2]David M. Routman, M.D. - routman.david@mayo.edu,**
**[1]Brady S. Laughlin, M.D. - laughlin.brady@mayo.edu,**
**[1]Carlos E. Vargas, M.D. - vargas.carlos@mayo.edu,**
**[1]Samir H. Patel, M.D. - patel.samir@mayo.edu,**
**[1]Sujay A. Vora, M.D. - vora.sujay@mayo.edu,**
**[2]Nadia N. Laack, M.D. - laack.nadia@mayo.edu,**
**[2]Andrew Y.K. Foong, Ph.D. - foong.andrew@mayo.edu,**
**^[1]Wei Liu, Ph.D. - liu.wei@mayo.edu,**
**^[2]Mark R. Waddle, M.D. - waddle.mark@mayo.edu**

***co-first authors**
**^co-corresponding authors**

**1. Department of Radiation Oncology, Mayo Clinic, Phoenix, AZ, United States**
**2. Department of Radiation Oncology, Mayo Clinic, Rochester, MN, United States**
**3. Division of Radiation Oncology, IEO, European Institute of Oncology, IRCCS, Milan, Italy**

**4. Department of Oncology and Hemato-Oncology, University of Milan, Milan, Italy**
**5. Department of Surgery, Mayo Clinic, Rochester, MN, United States**

**AI Acknowledgement:**

ChatGPT (OpenAI, GPT-5.2) was used to improve manuscript readability and language clarity and to assist in generating portions of figures included in the diagrams; all scientific content, data analysis, interpretations, and final figure design were independently reviewed, verified, and approved by the authors, who take full responsibility for the accuracy and integrity of the work.

**Abstract**

**Importance:** Radiation oncology workflows generate large volumes of heterogeneous electronic health record (EHR) data that clinicians must synthesize daily. This manual preparation contributes to inefficiency and cognitive burden. Large language models (LLMs) offer the potential to automate information synthesis, but evidence of real-world, workflow-embedded implementations at scale remains limited.

**Objective**: To describe the design and early clinical evaluation of The Daily Dose (TDD), an LLM-driven, automated clinical summarization and clinical-trial identification system integrated into routine radiation oncology practice.

**Design**: Mixed-methods evaluation using a cross-sectional, anonymous clinician survey administered after 1 month of system deployment.

**Setting**: Radiation oncology departments across 3 Mayo Clinic campuses (Arizona, Florida, and Minnesota).

**Participants**: All health workers receiving TDD during the survey period, including attending physicians, advanced practice providers, nurses, and trainees.

**Exposure**: Daily automated delivery of physician-specific email summaries generated using RadOnc-GPT, including patient schedules, concise EHR-derived clinical-status summaries, and automated identification of potentially relevant clinical trials for new or consult visits.

**Main Outcomes and Measures**: Primary outcomes included self-reported usability, satisfaction, perceived usefulness, perceived impact on workflow, time savings, and intention for continued use. Internal consistency reliability was assessed using Cronbach's $\alpha$.

**Results**: Among 55 respondents, 52 (94.5%) worked in radiation oncology, and 38 (69.1%) were attending physicians. Most participants (83.6%) reported using TDD daily or several times per week. Mean (SD) scores were 3.89 (1.04) for usability and satisfaction, 3.43 (1.24) for perceived usefulness, and 3.80 (1.17) for impact and future use (5-point Likert scale). Overall satisfaction was positively associated with perceived time savings ($p < .001$). Participants reported variable time savings, with 27% estimating ≥10 minutes saved per day. The questionnaire demonstrated excellent internal consistency (overall Cronbach's $\alpha$ = 0.97). Qualitative feedback highlighted improved preparedness and patient-context awareness, particularly when covering for colleagues, alongside concerns regarding occasional inaccuracies and imperfect clinical-trial matching.

**Conclusions and Relevance**: In this early evaluation, a workflow-integrated LLM-based automation tool was widely adopted and perceived as usable and valuable by radiation oncology clinicians, with heterogeneous but generally favorable assessments of its impact. These findings suggest that LLM-driven preparatory tools may enhance clinician readiness and situational awareness in complex care environments.

**List of Acronyms**

- AAPM: American Association of Physicists in Medicine
- AKI: Acute kidney injury
- APP: Advanced practice provider
- EHR: Electronic health record
- GPT: Generative Pre-trained Transformer
- HIPAA: Health Insurance Portability and Accountability Act
- IRB: Institutional Review Board
- LLM: Large language model
- MCQ: Multiple choice question
- NCCN: National Comprehensive Cancer Network
- NP: Nurse practitioner
- PA: Physician assistant
- PSA: Prostate-specific antigen
- RN: Registered nurse
- SUS: System Usability Scale
- TAM: Technology Acceptance Model
- TAES: TriAl Eligibility Surveillance (system)
- TDD: The Daily Dose
- TRIPOD-LLM: Transparent Reporting of a multivariable model for Individual Prognosis or Diagnosis–Large Language Models

## Introduction

Radiation oncology harnesses a uniquely complex digital ecosystem to deliver precise and personalized cancer treatment, with each patient's treatment course resulting in the generation of hundreds of data points spanning years. Electronic health records (EHRs) have standardized access to necessary medical information, but the level of information can become overwhelming and require radiation oncologists and staff to access multiple software programs to collect this information. This manual process contributes to inefficiency, cognitive fatigue, and reduced time for direct patient interaction.

Large language models (LLMs) have rapidly advanced the automation of information-intensive clinical tasks[1,2]. Fine-tuned transformer architectures such as LLaMA have been shown to generate discharge and hospital-course summaries and CT simulation summaries with accuracy comparable to professionally authored documentation while maintaining factuality and readability in HIPAA-compliant settings[3–7]. Multi-agent and retrieval-augmented frameworks have extended these capabilities to administrative and clinical operations, automating record retrieval, patient scheduling, documentation, and data integration across hospital systems[8,9]. Complementary studies have demonstrated that natural-language methods can streamline routine processes such as diagnostic and procedural coding[10] and that LLM-based feature selection can enhance predictive modeling in health-care data pipelines[11]. Collectively, these advances illustrate the growing capacity of LLMs to automate complex, language-based tasks in real health-care environments.

RadOnc-GPT was developed as a specialized LLM agent for radiation-oncology workflows, capable of outcome labeling, documentation assistance, cancer patient education[19], and inbox message generation with expert-level performance[20,21].

Leveraging these foundational works, we developed *The Daily Dose* (TDD)—a production-level application that operationalizes RadOnc-GPT within the live clinical workflow. Each morning, *The Daily Dose* automatically generates a personalized summary of each patient visit for every radiation oncologist at the start of the day, delivered via email. This concise, EHR-derived clinical-status summary also automatically identifies clinical trials for which patients may be eligible. *The Daily Dose* is now deployed across three Mayo Clinic campuses (Arizona, Florida, and Minnesota) and represents a first-in-practice implementation of LLM-driven automation in daily oncology care at scale. This manuscript describes the system's design and initial evaluation, focusing on clinician perceptions of usability, satisfaction, and perceived impact on workflow.

# Material and methods

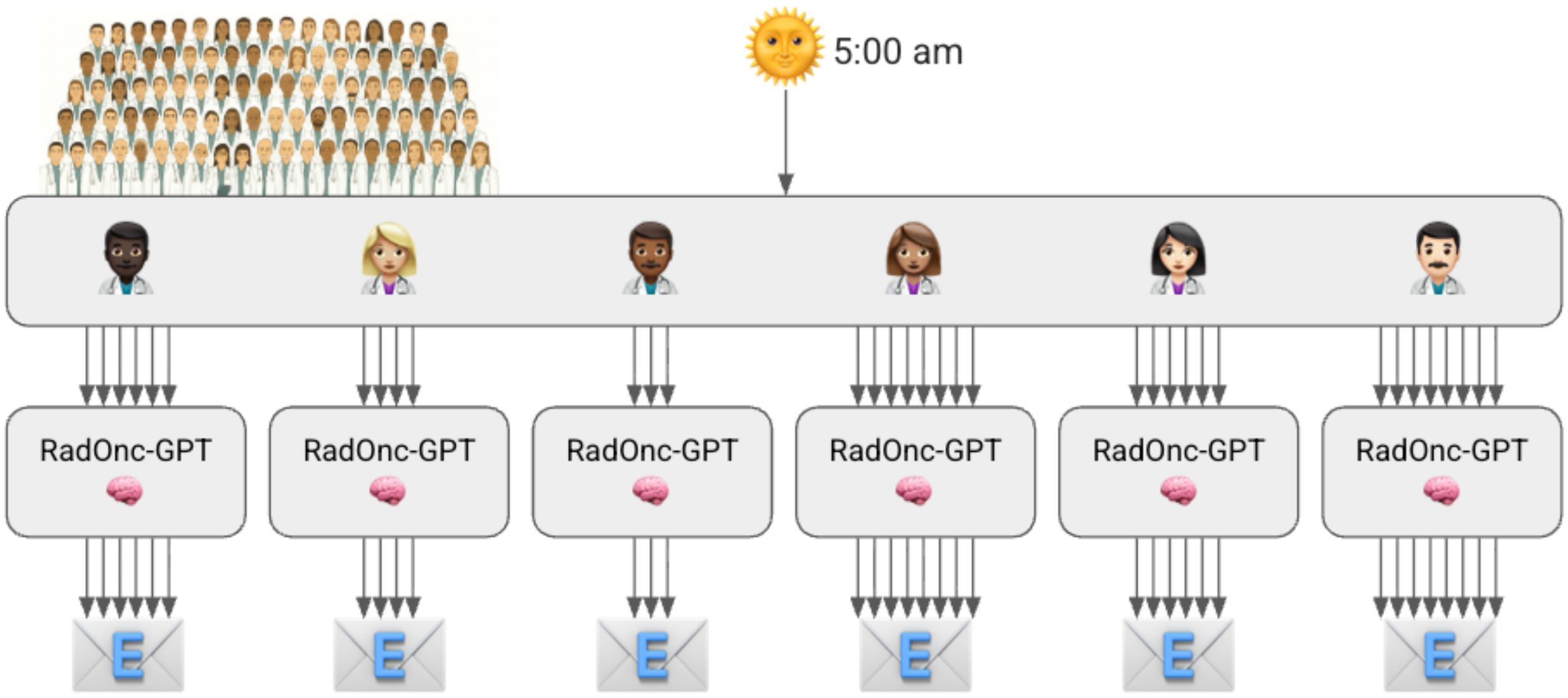

**Figure 1.** System Diagram for *The Daily Dose*. Each morning, RadOnc-GPT retrieves the Radiation Oncologist' schedule for the day and summarizes their patients. This summary is sent as a single email at the start of the day.

*The Daily Dose (TDD)* is a fully automated, provider-facing application that integrates directly with the departmental clinical workflow to deliver daily, physician-specific, AI-generated patient summaries by email. The system was designed to run autonomously within Mayo Clinic's secure computing environment, using RadOnc-GPT, an LLM-based agent tooled for radiation oncology, as its reasoning engine[20]. Over the course of this evaluation, the LLM used was GPT-4o (OpenAI, San Francisco, CA) with temperature set to 0. All processing occurs in a HIPAA-compliant environment.

*The Daily Dose* executes each morning at 5:00 AM local time. A coordinating script identifies all participating physicians and initiates TDD generation process for each in parallel.

### Clinical Summary

For every provider, the system retrieves the current appointment list from the EHR including the appointment time, type, patient name, and patient identifier. For each appointment, TDD prompts RadOnc-GPT to generate concise, context-aware clinical summaries. RadOnc-GPT is prompted to retrieve relevant EHR elements, including diagnosis details, treatment plans, appointments, radiology and pathology reports, and clinical notes, which are then synthesized into a short narrative describing the patient's current condition, with the context being that the provider is about to visit the patient. According to the prompting, disease-specific parameters (e.g., prostate cancer PSA trends, Gleason score, and NCCN risk category) are automatically included when relevant. When insufficient data are available, as determined by RadOnc-GPT, RadOnc-GPT outputs a standardized fallback statement: *"There is not enough information to provide a status report". The f*ull prompt used for generating clinical summaries is available in the Supplemental Material.

**Clinical Trial Matching**

For patients with *new* or *consult* visits, TDD further prompts RadOnc-GPT to evaluate the patient for potential clinical trial eligibility (full prompt in supplemental material). RadOnc-GPT is prompted to first identify the patient's primary diagnosis and relevant treatment context to generate condition-intervention keyword pairs, which are then used to iteratively search clinical-trial databases via the clinicaltrials.gov API with studies restricted to trials at Mayo Clinic that are open and actively accruing. Next, the study inclusion and exclusion criteria for each candidate trial are assessed, summarizing whether the patient likely meets, does not meet, or lacks sufficient information for each criterion. The final output lists potentially relevant trials with short eligibility summaries and study links. If no suitable trials are found or patient demographics are incomplete, standardized fallback text is inserted to ensure transparency.

**Email Format and Delivery**

Once all summaries are generated, TDD constructs an email tailored to each physician. The message includes:

1. A greeting and the physician's name,
2. Each scheduled appointment with appointment type and intuitive visual indicators (e.g., emojis to distinguish consults, follow-ups, and simulations),
3. The concise patient-status summary,
4. Clinical-trial eligibility results (for new or consult patients only), and
5. A closing section with contact information and feedback link.

All content is rendered from Markdown, sent through the institutional mail server, and archived automatically. Log files record generation time, model interactions, and any errors. Failed processes are retried up to three times to ensure completion. Physicians typically receive their personalized "Daily Dose" email before the clinical day begins, allowing rapid review of upcoming cases and potential matching clinical trials.

**Questionnaire and Feedback**

One month after launching TDD to all of radiation oncology at Mayo Clinic, a structured feedback questionnaire was sent to evaluate the usability and satisfaction with The Daily Dose. The survey was conducted on Microsoft Forms and was available from the 1st to the 27th of August 2025. A weekly reminder was sent by email to the users to complete the survey. Participation was voluntary, responses were anonymous, and informed consent for research purposes was implied by completion.

The questionnaire content was adapted from System Usability Scale (SUS)[22] and Technology Acceptance Model (TAM)[23]. Given the nature of *The Daily Dose*, items were modified to capture domains relevant to clinical workflow integration.

The final survey included 11 items divided into 5 domains:

1. Demographics: three multiple choice questions (MCQs) capturing the specialty, the professional role, and years in practice.
2. Usage patterns: two MCQs assessing frequency of reading summaries, estimation of the daily time saved.
3. Usability and satisfaction: five Likert-scale items, adapted from SUS, evaluating ease of use, clarity of summaries, ability to locate relevant information, integration into daily workflow and overall satisfaction.
4. Perceived usefulness, impact and future use: nine Likert-scale items, adapted from TAM, assessing the perceived usefulness and impact on clinical practice, the intention to continue use and likelihood of recommendation.
5. Optional comments: three open questions assessing the most useful parts of the tool, suggestions for improvements and additional comments.

**Statistical analysis**

Descriptive statistics included means, standard deviations, and frequency distributions for all variables. Likert scale responses (1-5 scale) were converted to numeric values and domain scores were calculated as the mean of constituent items within each domain. Internal consistency reliability was assessed using Cronbach's alpha coefficient for each domain and the overall scale.

Non-parametric statistical tests were employed due to the ordinal nature of Likert scale data and potential non-normal distributions. Spearman rank correlation coefficients, Kruskal-Wallis test and Mann-Whitney U tests were used when appropriate.

Statistical significance was set at $p < 0.05$ for all analyses. All analyses accounted for missing data through pairwise deletion, and results are presented with appropriate confidence intervals where applicable.

This project was reviewed and exempted by the Mayo Clinic Institutional Review Board, and conducted as a quality improvement initiative.

This study is reported in accordance with the TRIPOD-LLM reporting guideline for studies involving large language models[24] (Supplemental material).

## Results

A total of 55 out of 110 participants (50%) completed the survey. The vast majority (94.5%) worked in radiation oncology, with a small number representing medical oncology (3.6%) and oncology nurse navigation (1.8%). Most respondents were attending physicians (69.1%), followed by advanced practice providers (14.5%) and registered nurses (10.9%). More than half of the participants (56.4%) had over 10 years of professional experience. In terms of utilization patterns, 52.7% of respondents reported using TDD daily, and 30.9% used it a few times per week. **Table 1**.

Perceived time savings varied across users. While 7 (12.7%) participants reported saving more than 20 minutes per day and 8 (14.5%) reported 10–20 minutes, 10 (18.2%) indicated saving 5–10 minutes, and 18 (32.7%) reported saving less than 5 minutes. Conversely, 12 (21.8%) respondents stated that they did not experience any time savings when using TDD.

| Total | 55 (100%) |
| --- | --- |
| *Specialty* | |
| Radiation oncology | 52 (94.5%) |
| Medical oncology | 2 (3.6%) |
| Oncology nurse navigator | 1 (1.8%) |
| *Role* | |
| Attending physician | 38 (69.1%) |
| APP (PA or NP) | 8 (14.5%) |
| RN | 6 (10.9%) |
| Resident / Fellow | 2 (3.6%) |
| LPN | 1 (1.8%) |
| *Years of Experience* | |
| >10 years | 31 (56.4%) |
| 5-10 years | 12 (21.8%) |
| <5 years | 12 (21.8%) |
| *Usage Frequency* | |
| Daily | 29 (52.7%) |
| A few times a week | 17 (30.9%) |
| Occasionally | 5 (9.1%) |
| Rarely or never | 4 (7.3%) |

**Table 1**. Characteristics of the survey respondents.

*Usability and Satisfaction*

The usability and satisfaction section yielded a mean score of 3.89 (±1.04). The majority of the respondents (87%) agreed that TDD was easy to use. Similarly, 76% of participants agreed that the summaries are clearly structured and easy to read. Over half of participants (56%) declared that they could quickly locate the information they need, and 60% reported that the tool integrated well with their daily workflow. In terms of overall satisfaction with the performance of TDD, 7% of the respondents strongly disagreed, 24% disagreed, 20% were neutral, 16% agreed and 33% strongly agreed. **Figure 2a**.

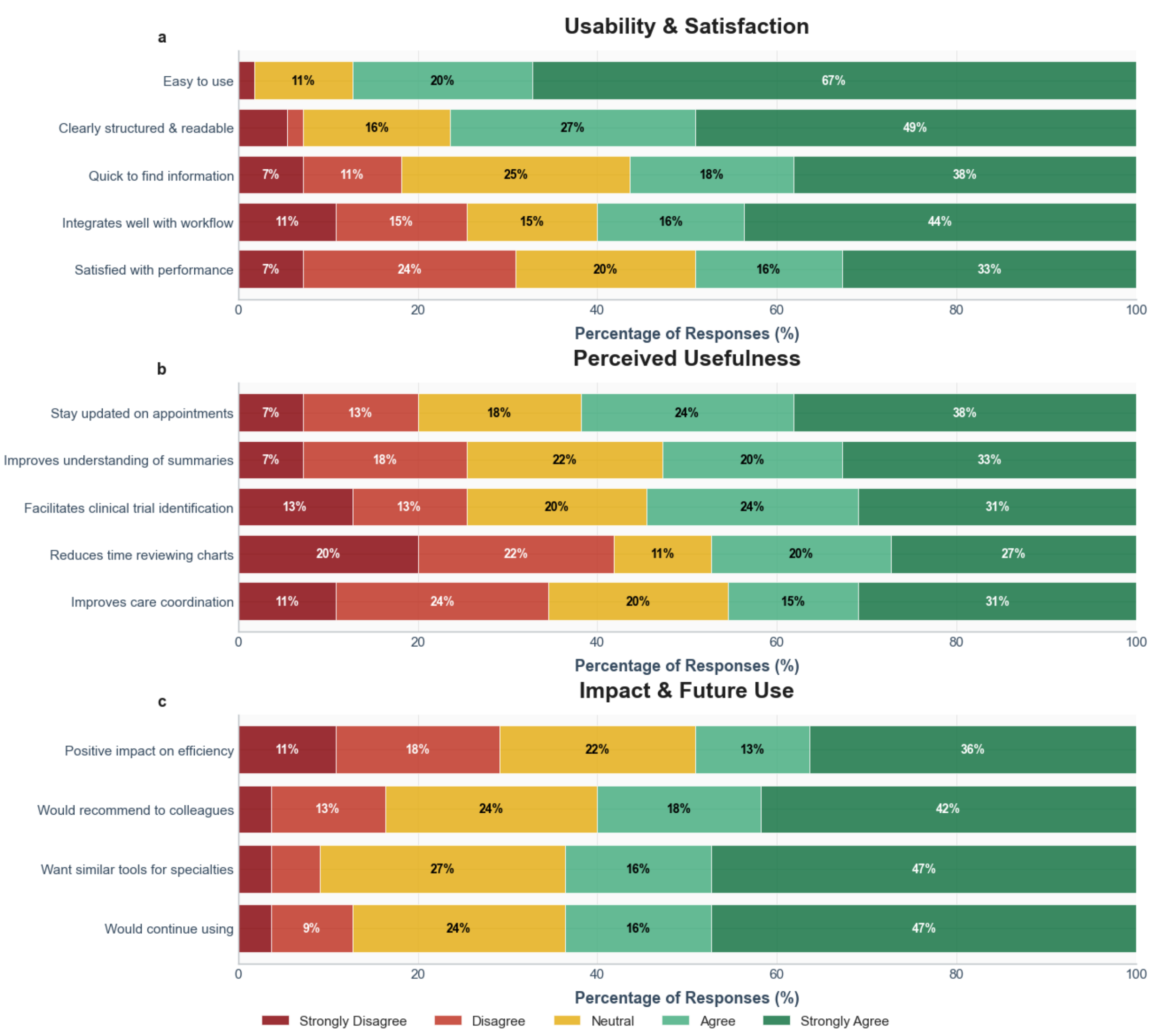

**Figure 2**. Likert scale ratings of a) usability and satisfaction; b) perceived usefulness ratings; c) impact and future use ratings.

*Perceived usefulness*

The perceived usefulness section received a mean score of 3.43 (±1.24). The 62% of the respondents agreed or strongly agreed that TDD helped them stay updated on scheduled appointments, and 53% agreed that it improved the understanding of patient clinical summaries. Similarly, 55% indicated that the tool facilitated the identification of patients eligible for clinical trials. In contrast, responses were more divided regarding TDD's contribution to team care coordination, with 35% disagreeing and 20% remaining neutral. Likewise, perceptions were mixed when asked if TDD was helping reduce the time spent on reviewing charts or preparing rounds, with 20% of respondents strongly disagreeing, 22% disagreeing and 11% being neutral. **Figure 2b**.

*Impact and future use*

The impact and future use score yielded a mean score of 3.80 (±1.17). A majority (64%) of the respondents agreed or strongly agreed that they would like to see similar tools implemented in other specialties and that they would continue using TDD, if it remains available. Additionally, 60% of participants would recommend the use of the tool to a colleague, while 17% expressed disagreement. In contrast, responses were mixed when asked if TDD has a positive impact on clinical efficiency, with 49% of respondents agreeing or strongly agreeing and 29% disagreeing or strongly disagreeing. **Figure 2c**.

*Overall satisfaction*

The overall satisfaction analysis yielded a mean score of 3.70 ± 1.10. **Figure 3** illustrates the relationship between overall satisfaction and perceived time saved when using TDD. A significant difference in satisfaction levels was observed across time-saved categories ($p < 0.001$), with respondents reporting greater satisfaction as the amount of time saved increased. Participants who indicated saving more than 5 minutes demonstrated the highest satisfaction scores, while those who reported no time savings expressed the lowest satisfaction levels.

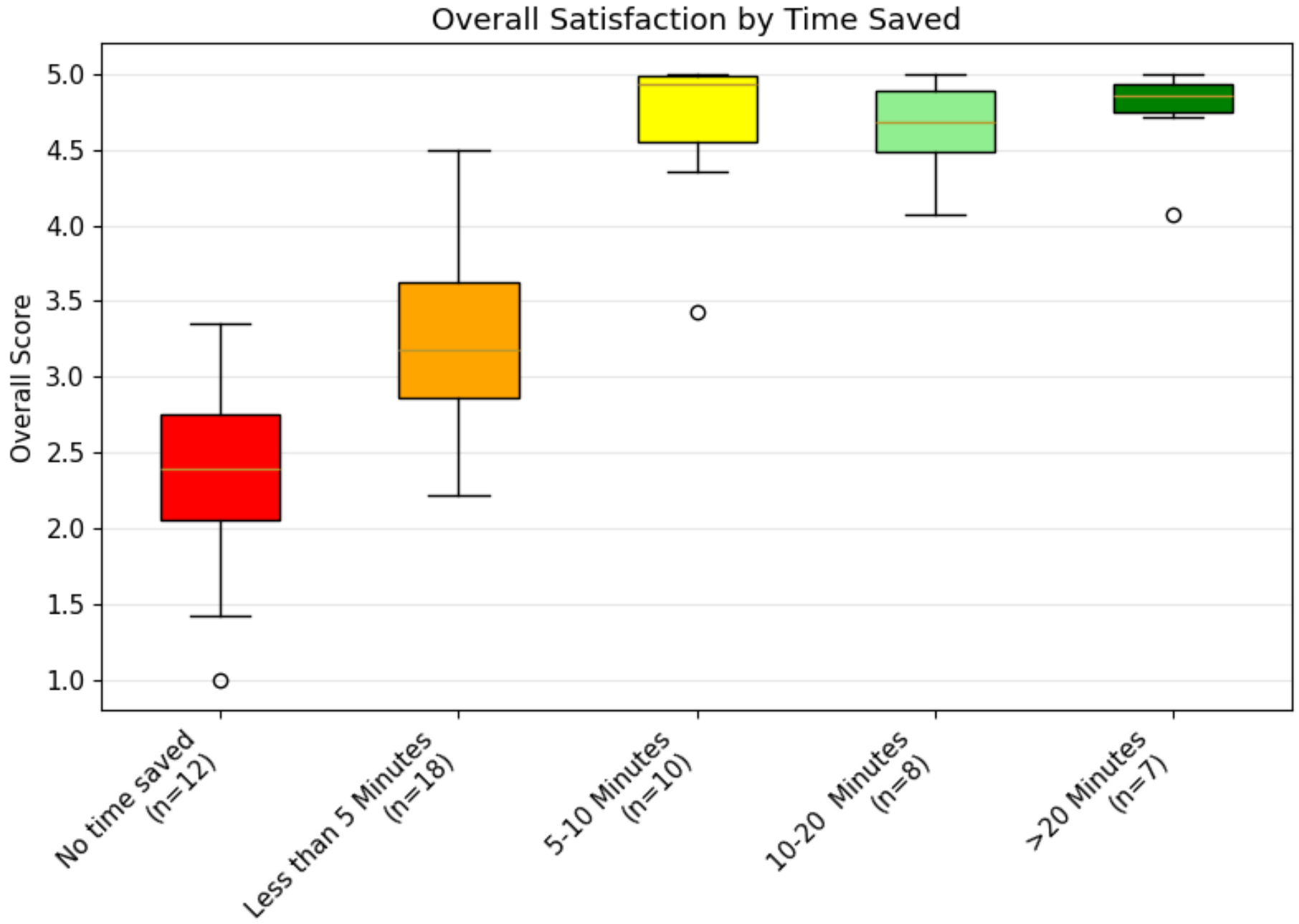

**Figure 3**. Overall Satisfaction by Time Saved When Using The Daily Dose.

The breakdown of overall satisfaction by seniority level (**Figure 4**) showed mean scores of 3.42 ± 0.82 for staff with <5 years of experience, 3.81 ± 0.79 for those with 5-10 years of seniority, and 3.77 ± 1.29 for staff with >10 years of experience (p = 0.607).

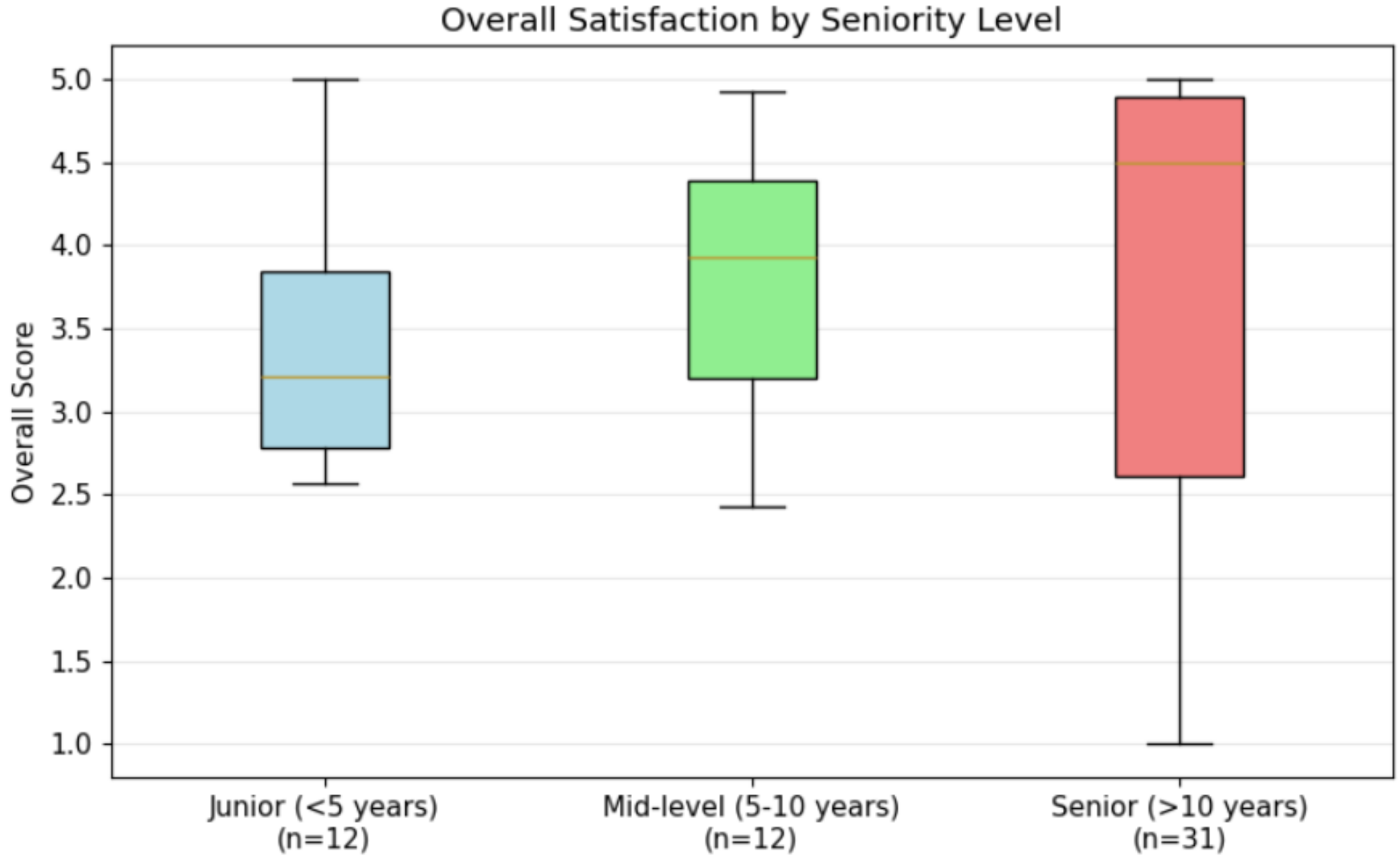

**Figure 4**. Overall Satisfaction by Seniority level

*Internal Consistency*

The internal consistency of the questionnaire was excellent, with an overall Cronbach's alpha of 0.971, indicating high reliability of the scale. Subscale analyses also demonstrated strong internal consistency, with Cronbach's alpha values of 0.906 for the Usability and Satisfaction section, 0.937 for the Perceived Usefulness section, and 0.951 for the Impact and Future Use section.

*Free-text feedback*

In the open question section, responders particularly highlighted the opportunity of receiving concise and helpful summaries of patient information, which offered a better understanding of patient history and treatment trajectory, and aided in daily preparation, especially when covering for other colleagues. However, areas for improvement were also noted. Some users reported that summaries could be incorrect, include unnecessary details, or miss critically important information, indicating a need for enhanced access to outside records and improvements in accuracy and relevance. Additionally, while the identification of potential clinical trials is a valued feature, some feedback suggested that the system frequently identifies erroneous trials or suggestions that do not meet inclusion/exclusion criteria, requiring further refinement.

## Discussion

*Overview and adoption*

*The Daily Dose* represents one of the first LLM-based automation tools to be deployed across all providers within a single specialty at a large institution, spanning three Mayo Clinic regions and embedded directly into the routine clinical workflow. One-month post-deployment questionnaire responses were heterogeneous across all domains, however the majority of users reported agreement or strong agreement for key domains such as usability, time savings, and satisfaction and more than half of users were daily users. These results indicate an overall favorable, but somewhat fragmented, perception of *The Daily Dose,* which is consistent with the expectation for an early-phase tool which has been deployed in a real-world scenario. The high adoption rate suggests that the tool provides sufficient perceived value to be incorporated into daily routines, meanwhile the high usability scores and the strong intention for continued use support the low interaction burden and the seamless integration into the existing clinical workflows.

*Time savings, satisfaction, and preparedness*

Perceived time savings varied substantially among individual users. Using the midpoint of each reported time-saved, the cumulative perceived time savings across the entire cohort of respondents was approximately **560 minutes per day**. This estimate should be interpreted cautiously, since it relies on self-reported intervals and on the assumptions of category midpoints. Nonetheless, it suggests that even modest individual gains may translate into a meaningful collective efficiency benefit when scaled across a larger clinical group.

Interestingly, the perceived time savings varied widely and did not uniformly translate into reduced chart-review time. This discrepancy may reflect the role of *The Daily Dose* as a preparatory and organizational aid rather than a replacement for clinical decision-making or formal documentation review. In this context, the finding that half of respondents agreed or strongly agreed with the key usefulness and impact statements should be considered a meaningful signal, rather than a limitation. However, the association between satisfaction and perceived time savings suggests that efficiency gains, when realized, influence the user experience meaningfully, consistent with the pattern observed in Figure 4 where overall satisfaction increased with greater time saved.

*Clinical trials: context and implementation impact*

One domain in which LLMs have demonstrated significant promise is clinical-trial matching, a traditionally labor-intensive process that requires manual review of patient records and protocol criteria. Early cognitive-computing approaches such as IBM Watson for Clinical Trial Matching at Mayo Clinic and the TriAl Eligibility Surveillance (TAES) system established feasibility for automated eligibility detection within EHRs[25,26], and subsequent institution-agnostic frameworks expanded matching capabilities across networks[27]. More recent LLM-driven systems, including TrialGPT and other multimodal or retrieval-augmented pipelines, have reported criterion-level accuracies in the range of approximately 85–93% and reductions in manual screening time of more than 40%[28–31]. While those studies largely emphasize technical performance, our implementation differs in that The Daily Dose is evaluated as a workflow-embedded intervention delivered daily to clinicians at scale. In that context, the value is not only in perfect eligibility determination, but also in supporting routine consideration of trials within daily planning.

Consistent with this framing, free-text responses suggested an additional and somewhat unexpected effect: the tool helped at least some physicians develop a better understanding and awareness of ongoing clinical trials, particularly those relevant to their specialty. This educational impact may be meaningful even when trial suggestions require verification. At the same time, open-question feedback also highlighted concerns regarding occasional inaccuracies or irrelevant trial suggestions, showing the need for further refinements to ensure a sustained clinical trust over time.

*Measurement properties and subgroup patterns*

The questionnaire demonstrated an excellent internal consistency across domains, supporting the coherence of the grouped items. Responses spanned the full Likert scale, suggesting that the questionnaire was able to capture meaningful differences in response patterns across different professional roles. When the differences in perceived benefit across seniority levels were analyzed, senior responders were more prone to score higher the tool in comparison to younger colleagues, however, these trends were not statistically significant and should be viewed as hypothesis generating (see Figure 5).

*Limitations and future work*

Several limitations of the questionnaire should be acknowledged. First, the survey captured user perceptions rather than objective workflow metrics, so responses may reflect individual expectations and prior experience with digital tools. Second, although adapted from established frameworks, the instrument was not formally validated as a standalone scale, and results should therefore be interpreted as exploratory. Third, participation was voluntary, introducing potential selection bias if users with stronger positive or negative views were more likely to respond. Finally, the timing of the questionnaire, limited sample size, and predominance of radiation oncologists may reduce generalizability to other settings.

A key limitation affecting new-patient summarization and clinical-trial matching is that RadOnc-GPT did not retrieve external records. Because performance depends on input completeness, sparse clinical context can limit reliable eligibility assessment, whereas richer longitudinal data should improve trial identification and criterion reasoning. Future work will prioritize integrating external records to increase data completeness.

Scalability is also important: current costs are relatively high due to extensive EHR retrieval/summarization and, for consults, trial-eligibility evaluation. To reduce costs, we will improve data selection to minimize unnecessary retrieval and token use and selectively use lower-cost LLMs, reserving more capable models for complex or ambiguous cases. Future evaluations should incorporate objective workflow metrics and formal assessments of summary accuracy and omission rates to better quantify clinical impact and guide iterative improvements.

## Conclusion

*The Daily Dose* has emerged as a valuable clinical tool for radiation oncology practice at Mayo Clinic in the modern era of complex patients, fragmented digital infrastructure, and time constraints. Based on the results of this study, *The Daily Dose 2.0* has been developed which provides access to outside records (via EHR connection or PDF documents), a more selective and lower costs two stage data retrieval architecture, a largely redesigned clinical trial matching system, and visit specific prompt databases that will allow roll-out to every physician in the Mayo Clinic Comprehensive Cancer Center.

## Supplemental Material

### *Clinical Summary Prompt*

Patient {physician_appointment["patient id"]} has one or more appointments today. You will find instructions below on how to write a brief summary of the current status of the patient, which will be read by their radiation oncologist (provider) just prior to their visit. Follow these steps without pausing, i.e. DO NOT issue a `<DONE>` signal until AFTER you complete all the steps and output the JSON summary inside a code block:

Step 1: Retrieve the following patient data simultaneously: patient details, patient treatment details, patient diagnosis details, patient appointments (today's appointments only), patient radiology reports, patient pathology reports, and patient clinical notes (radiology, pathology, surgery, medical oncology, ENT, urology, and radiation oncology).

Step 2: Summarize important events in the form of a timeline (this step will help to prepare you to write a good patient status summary).

Step 3: Write a brief overall status that includes their current diagnosis, important treatment information and dates relating to their current diagnosis (including but not limited to the number of fractions received/prescribed, dose received/prescribed, treatment start date, most recent treatment date, and the next treatment date - if there are delivered treatment details), a brief mention of prior radiation treatments in their history - not the current treatment (e.g. 75 Gy in 2022, 48 Gy in 2017, etc.), medications the patient is currently taking (especially relating to the cancer treatment - also include duration), important recent events, any ongoing complications or discomfort, recent imaging results (including dates), the reason for today's visit (can be found by in the appointments information), other appointments the patient has today, and anything else you think the radiation oncologist needs to know just prior to visiting with the patient. If today's appointments include a management visit, then you need to mention anything that was attempted in the prior visit (see the most recent progress note) in order to improve the patient's condition. This way, the oncologist will remember to ask about it. Your summary should be a rich, yet concise paragraph without formatting, something that quickly provides the information to the physician. Also, note the following disease-site-specific requirements: if prostate cancer, note the highest Gleason score on biopsy, the most recent PSA, the PSA closest to the diagnosis onset date, the risk category per NCCN (low risk: T1-T2a, Gleason score <=6, and PSA <10 ng/mL; intermediate risk: T2b-T2c or Gleason score of 7 or PSA of 10-20 ng/mL; high risk: T3a or Gleason score >=8 or PSA >20 ng/mL; very high risk: T3b-T4 or primary Gleason pattern 5 or >5 cores with Gleason score >=8 and metastatic N1 or M1 with any T stage), and if intermediate risk - whether it is favorable (only one risk factor for intermediate risk and <50% of biopsy cores positive) or unfavorable (2 or more risk factors, primary Gleason 4, or 50% or more biopsy cores positive). In order to allow for parsing, provide your summary in JSON format within a code block as follows (the JSON still needs to be with the <CHAT> scope):

```
{{"patient_status_summary": "This is where a brief summary of the patient status that will help the physician with their patient visit goes."}}
```

If no patient data could be retrieved, then return the following:

```
{{"patient_status_summary": "There is not enough information to provide a status report."}}
```

# *Clinical Trial Evaluation Prompt*

Your task is to determine which clinical trials patient {physician_appointment["patient id"]} might be eligible to participate in. To do this, you will first retrieve the patient data and build a timeline of important events. Next, you will determine a set of potential clinical trials to evaluate. Finally, you will evaluate each potential clinical trial one by one. Follow the detailed steps below in order to guide you through this process. Note that you should perform every step without pausing, i.e. do not issue a <DONE> signal until after the Results Output.

---
## **Build A Patient Timeline Table**

1. **Retrieve Patient Data**
   - Retrieve (simultaneously) and then briefly summarize the following patient data (using the <CHAT> scope):
     - patient details
     - patient diagnosis details
     - patient appointments
     - pathology reports
     - radiology reports
     - pathology clinical notes
     - radiology clinical notes
     - surgery clinical notes
     - medonc clinical notes
     - radiation oncology clinical notes
     - urology clinical notes
     - ENT clinical notes

2. **Construct The Patient Timeline**
   - Build a timeline of important clinical events (e.g. diagnostic tests or imaging, lab results, diagnosis, staging, ECOG or Karnofsky performance scores, current symptoms or side effects, significant comorbidities or conditions that commonly impact eligibility such as diabetes/autoimmune diseases/cardiac events, biomarker and genetic testing results, simulation, prior surgeries, treatment planning, turning points in care, etc.).
   - Write out the timeline as a table using Markdown. The timeline will help towards building a coherent overview of the patient that can better inform keyword generation and more accurate trial exclusion decisions later.

---
## **Keyword Generation**

1. **Restate The Patient's Condition**
   - Briefly outline the condition(s) that you are going to focus on for clinical trial matching (these are likely near the end of the timeline, i.e. recent).

2. **Identify Core Condition Keywords**
   - List single-term or short-phrase descriptors that best capture the condition.
   - Have a detailed discussion about the relevancy of the conditions you provided (using the <CHAT> scope). After the discussion, rank the conditions by relevancy.

3. **Determine Potential Intervention Keywords**
   - List possible interventions, treatments, therapies, adjuvant therapies, generic drugs, etc. that might be used to treat the conditions you listed.

- If breast cancer, you might consider these terms in addition to others: aromatase inhibitors, SERMs, SERDs, ovarian suppression, and progesterone therapies (these are major classes of endocrine therapy).
- Have a detailed discussion about the relevancy of the interventions you provided (using the <CHAT> scope). After the discussion, rank the interventions by relevancy.

4. **Generate Multiple Condition Keyword And Intervention Keyword Sets For Search**
- Write out at least 10 different condition-intervention keyword combinations (Make it clear what are the conditions and what are the interventions).
- Have a detailed discussion about which combinations are most relevant (using the <CHAT> scope). After having a discussion, rank the combinations by relevancy.

---
## **Synonym Expansion**

1. **Generate Synonymns For The Conditions And Interventions**
- Now that we have a highly relevant list of conditions-intervention combinations, we want to make sure our search captures the intented results for each condition-intervention combination. To do this, you should replace each keyword by itself plus its synonyms. Working from the ranked list of relevant condition keyword and intervention keyword sets, replace each keyword by itself and its synonyms.
- Now translate the list into the exact inputs you'll use in the `get_list_of_clinical_trials` function. There should be a separate set of inputs one to one with the list of ranked conditions-interventions combinations. List them in the order that you will process them (as part of the iterative search process).

## **Building The Clinical Trials Pool With Iterative Searching**

1. **Dynamic (Iterative) Search**
- Perform a search for clinical trials using the search inputs you came up with (and the age and sex of the patient) starting from the most relevant search input. Indicate how many trials were found. We want our pool of potential clinical trials to contain something like 7 to 15 trials, so you may need to conduct multiple searches using the next-most relevant search inputs in the list your previously came up with, one by one, until something like 7 to 15 relevant trials are found in the end. After each search, indicate how many unique trials have been found so far (cumulative value - make sure to not include duplicates as part of the accounting). You must perform a minimum of two searches in this process, but no more than 5. If you end up with less than 7 trials after 5 searches, just evaluate them as-is without conducting another search. If you aren't getting any results, you may need to combine more keywords as a last ditch attempt.
- After iteratively searching, this final set of potential clinical trials represents the pool of trials to evaluate.

---
## **Exclusion Process**

1. **Evaluate Detailed Eligibility Criteria**
- For the pool of trials that were found, look up each trial's eligibility requirements in detail.
- Summarize each criterion and note whether the patient meets it (met), does not meet it (not met), or if it is unknown.
- Provide supporting evidence from the patient data.

2. **Iterate Through All Relevant Trials**
- Perform the above eligibility-evaluation process (Step 1) for each remaining trial one by one.

- Continue until all relevant trials have been evaluated.

3. **Results Output**
    - Your final response, consisting of a summary of trials that were not excluded, must be formatted within an <ANALYSIS_SUMMARY> scope so that the user can extract the results.
    - It is possible that something may go wrong during your attempt to find which trials the patient might be eligible to participate in. The following notes provide multiple output scenarios.
    - If you were successful in finding some trials that the patient might be eligible to participate in, format your response as Markdown in a code block, wrapped in an <ANALYSIS_SUMMARY> scope surrounded with triple back-ticks (```), like this:

```<ANALYSIS_SUMMARY>
### Clinical Trials Eligibility Summary for {physician_appointment["patient name"]}
{{patient first name}} {{patient last name}} is potentially eligible to participate in the following clinical trials:

1. **NCTXXXXX:**
    - **Title:** Title of the first clinical trial
    - **Criteria Evaluation Summary:**
        - **Met:** summary of key criteria that were met...
        - **Unknown:** summary of criteria where it was unknown whether they were met or not...
        - **Not Applicable:** summary of criteria that were not applicable...
    - **URL:** A link to the clinical trial
2. **NCTXXXXXXX**:
    - **Title:** Title of the second clinical trial
    - **Criteria Evaluation Summary:**
        - **Met:** summary of key criteria that were met...
        - **Unknown:** summary of criteria where it was unknown whether they were met or not...
        - **Not Applicable:** summary of criteria that were not applicable...
    - **URL:** A link to the clinical trial
3. ...
</ANALYSIS_SUMMARY>```

    - If no trials were found, then return this:

```<ANALYSIS_SUMMARY>
### Clinical Trials Eligibility Summary for {physician_appointment["patient name"]}
No relevant clinical trials were found for {{patient first name}} {{patient last name}}.
</ANALYSIS_SUMMARY>```

    - If you coudn't determine the patient's age or sex, then return this:

```<ANALYSIS_SUMMARY>
### Clinical Trials Eligibility Summary for {physician_appointment["patient name"]}
Clinical trial eligibility could not be evaluated for patient {physician_appointment["patient id"]} because their age and sex could not be retrieved.
</ANALYSIS_SUMMARY>```

    - If an error occurs when searching for clinical trials, then return this:

```<ANALYSIS_SUMMARY>
### Clinical Trials Eligibility Summary for {physician_appointment["patient name"]}
An error occurred when searching for clinical trials for {{patient first name}} {{patient last name}}.

</ANALYSIS_SUMMARY>```